\documentclass{article}
\usepackage{ijcai22}

\usepackage{times}
\usepackage{soul}
\usepackage{url}
\usepackage[hidelinks]{hyperref}
\usepackage[utf8]{inputenc}
\usepackage[small]{caption}
\usepackage{graphicx}
\usepackage{amsmath}
\usepackage{amsthm}
\usepackage{booktabs}
\usepackage{algorithm}
\usepackage{algorithmic}
\title{Using contextual sentence analysis models to recognize ESG concepts}

\author{
Elvys Linhares Pontes$^1$
\and
Mohamed Benjannet$^1$\and
Jose G. Moreno$^2$\And
Antoine Doucet$^3$
\affiliations
$^1$Trading Central Labs, Trading Central, Paris, France\\
$^2$IRIT, UMR 5505 CNRS, University of Toulouse, Toulouse, France\\
$^3$L3i, La Rochelle Université, La Rochelle, France
\emails
elvys.linharespontes@tradingcentral.com 
}

\begin{document}

\maketitle

\begin{abstract}
This paper summarizes the joint participation of the Trading Central Labs and the L3i laboratory of the University of La Rochelle on both sub-tasks of the \textit{Shared Task FinSim-4} evaluation campaign. The first sub-task aims to enrich the ‘Fortia ESG taxonomy’ with new lexicon entries while the second one aims to classify sentences to either  ‘sustainable’ or ‘unsustainable’ with respect to ESG (Environment, Social and Governance) related factors.
For the first sub-task, we proposed a model based on pre-trained Sentence-BERT models to project sentences and concepts in a common space in order to better represent ESG concepts. 
The official task results show that our system yields a significant performance improvement compared to the baseline and outperforms all other submissions on the first sub-task.
For the second sub-task, we combine the RoBERTa model with a feed-forward multi-layer perceptron in order to extract the context of sentences and classify them. 
Our model achieved high accuracy scores (over 92\%) and was ranked among the top 5 systems.
\end{abstract}

\section{Introduction}
Financial markets and investors can support the transition to a more sustainable economy by promoting investments in companies complying to ESG (Environment, Social and Governance) rules. Today there is growing interest among investors in the performances of firms in terms of sustainability. Therefore, the automatic identification and extraction of relevant information regarding companies' strategy in terms of ESG is important.
The use of NLP (Natural Language Processing) methods adapted to the field of finance and ESG could help identify and process related information.

Taxonomies are important NLP resources, especially for semantic analysis tasks and similarity measures\cite{vijaymeena2016survey,bordea2016semeval}.
In this context, the FinSim4-ESG Shared Task proposed the tasks of enrichment of ESG taxonomy and sentences classification. FinSim-4 is the fourth edition of a set of evaluation campaigns that aggregate efforts on text-based needs for the Financial domain \cite{maarouf-etal-2020-finsim,Mansar-etal-2021-finsim2,kang-etal-2021-finsim3}. This latest edition is particularly challenging due to the continuously evolving nature of terminology in the domain-specific language of the ESG which leads to a poor generalization of pre-trained word and sentence embeddings.

Several studies addressed the problem of taxonomy generation for different domains \cite{shen2020taxoexpan,karamanolakis2020txtract}.
Deep learning based embedding networks, such as BERT \cite{devlin2018bert} have proven to be efficient for many NLP tasks. Malaviya \textit{et al.}~\shortcite{malaviya2020commonsense} used BERT for knowledge base completion and showed that BERT performs well for this task.
Liu \textit{et al.}~\shortcite{liu2020concept} used BERT to complete an ontology by inserting a new concept with the right relation.
Kalyan and Sangeetha~\shortcite{kalyan2021hybrid} used sentence BERT \cite{reimers-2019-sentence-bert} to measure semantic relatedness in biomedical concepts and showed that sentence BERT outperforms corresponding BERT models. Shen \textit{et al.}~\shortcite{shen2020biomedical} used sentence BERT to build a knowledge graph for the biomedical domain and showed that it obtains the best results.

For the Shared Task FinSim-4, we proposed several strategies based on BERT language models. For the first sub-task, we proposed a model based on pre-trained Sentence-BERT models to project sentences and concepts in a common space in order to better represent ESG concepts. 
For the second sub-task, we combined the RoBERTa model with a feed-forward multi-layer perceptron to extract the context of sentences and classify them. Official results of our participation show the effectiveness of our models over the Shared Task FinSim-4 benchmark. In terms of accuracy, our best runs respectively ranked 1$^{st}$ and 4$^{th}$ for the sub-tasks 1 and 2 with scores $0.848$ and $0.927$, respectively. 

The remainder of this paper is organized as follows. In Section 2, we present the shared task FinSim-4 and the datasets for both sub-tasks. Our proposed
models are detailed in Section 3. The setup and official results are described in Section 4. Finally, Section 5 concludes this paper.

\section{Shared Task FinSim-4}

The FinSim 2022 shared task aims to spark interest from communities in NLP, ML/AI, Knowledge Engineering and Financial document processing. Going beyond the mere representation of words is a key step to industrial applications that make use of natural language processing. The 2022 edition proposes two sub-tasks.

\subsection{Sub-task 1: ESG taxonomy extension}

The first sub-task aims to extend the `Fortia ESG taxonomy' provided by the organizers. This taxonomy was built based on different financial data providers' taxonomies as well as several sustainability and annual reports. It has twenty five different ESG concepts that belong to the ESG, split as: environment, social or governance. The organizers provide a training set which consists of terms belonging to each concept. This training set is unbalanced as one can observe in Table~\ref{tb:taxonomy-dataset} where one can find the number of terms for each concept in the train set.

Participants were asked to complete this taxonomy to cover the rest of the terms of the original `Fortia ESG taxonomy'. For example, given a set of terms related to the concept `Waste management' (e.g. Hazardous Waste, Waste Reduction Initiatives), participating systems had to automatically assign to it all other adequate 
terms. 

\begin{table}[tb]
\centering
\begin{tabular}{p{6cm}c}
\hline
\textbf{Concepts} & \textbf{\#terms} \\
\hline
Audit Oversight & 7 \\ 
Biodiversity & 29 \\ 
Board Independence & 2 \\ 
Board Make-Up & 37 \\ 
Carbon factor & 19 \\ 
circular economy & 47 \\ 
Community & 27 \\ 
Emissions & 39 \\ 
Employee development & 22 \\ 
Employee engagement & 23 \\ 
Energy efficiency and renewable energy & 59 \\ 
Executive compensation & 32 \\ 
Future of work & 18 \\ 
Human Rights & 10 \\ 
Injury frequency rate & 2 \\ 
Injury frequency rate for subcontracted labour & 35 \\ 
Product Responsibility & 51 \\ 
Recruiting and retaining employees (incl. work-life balance) & 11 \\ 
Share capital & 2 \\ 
Shareholder rights & 38 \\ 
Sustainable Food \& Agriculture & 54 \\ 
Sustainable Transport & 46 \\ 
Waste management & 16 \\ 
Water \& waste-water management & 21 \\ 
\hline
\end{tabular}
\caption{Dataset description for the ESG taxonomy extension sub-task.}
\label{tb:taxonomy-dataset}
\end{table}

\subsection{Sub-task 2: Sustainability classification}

The second sub-task aims to automatically classify sentences into sustainable or unsustainable sentences. A sentence is considered as sustainable if it semantically mentions the Environmental or Social or Governance related factors as defined in the Fortia ESG taxonomy. Table~\ref{tb:sustainability-dataset} summarizes the training data provided by the organizers.

\begin{table}[tb]
\centering
\begin{tabular}{lc}
\hline
\textbf{Classes} & \textbf{\#sentences} \\
\hline
Sustainable      & 1223  \\
Unsustainable    & 1042  \\
\hline
\end{tabular}
\caption{Dataset description for the sustainability sub-task.}
\label{tb:sustainability-dataset}
\end{table}

\section{Proposed strategies}
\subsection{Sub-task 1: ESG taxonomy extension}
\label{sc:approach1}

Semantic text similarity is an important task in natural language processing applications such as information retrieval, classification, extraction, question answering and plagiarism detection. This task consists in measuring the degree of similarity between two texts and to determine whether how semantically close they are (from completely independent to fully equivalent). In our case, the terms of a same concept are considered semantically equivalent. Siamese models have been shown to be effective on the semantic analysis of sentences~\cite{linhares-pontes-etal-2018-predicting,reimers-2019-sentence-bert}.

\begin{figure}[tb]
\includegraphics[width=6cm]{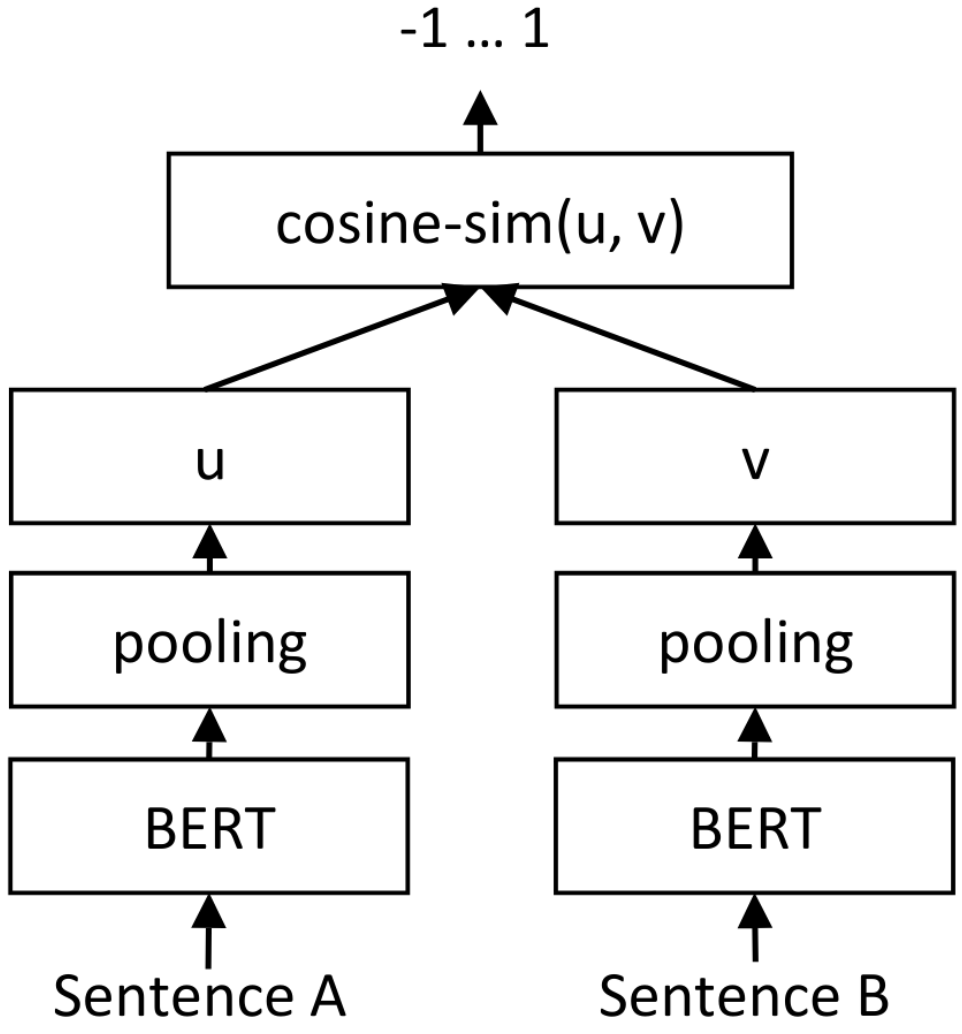}
\centering
\caption{Sentence transformer architecture at inference to compute semantic similarity scores between two sentences.}
\label{fig:sbert}
\end{figure}

Our model is based on Sentence-BERT (SBERT)~\cite{reimers-2019-sentence-bert}, a modification of the pre-trained BERT network that uses siamese and triplet network structures to derive semantically meaningful sentence embeddings that can be compared using cosine-similarity (Figure~\ref{fig:sbert}). This model is trained on a parallel dataset where two paraphrases or similar semantic sentences have high cosine similarity.

We consider all terms about a concept as paraphrases because they share the same semantic information. For instance, the terms `carbon footprint' and `carbon data' should have similar sentence representation because they share the same concept `carbon factor'; meanwhile, the terms `Water Risk Assessment' and `Transition to a circular economy' do not share the same concept and, consequently, their representations should have different sentence representation. 

With the SBERT model, we project all terms on the same dimensional space and then, we train our logistic regression model\footnote{\url{https://scikit-learn.org/stable/modules/generated/sklearn.linear_model.LogisticRegression.html}} to analyze and classify them to their corresponding concept classes.

\subsection{Sub-task 2: Sustainability classification}
\label{sc:approach2}

\begin{figure}[tb]
\includegraphics[width=6cm]{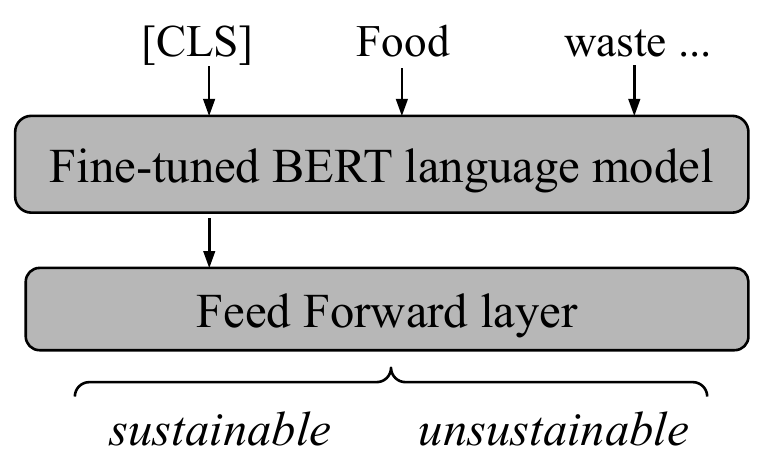}
\centering
\caption{Architecture model for the sustainability classification task.}
\label{fig:sustainability}
\end{figure}

For this sub-task, we combine a BERT-based language model~\cite{liu2019roberta} with a feed-forward multi-layer perceptron to extract the context of sentences and classify them into `sustainable' or `unsustainable'. The architecture of our model is described in Figure~\ref{fig:sustainability}.

We took the representation of the [CLS] token at the last layer of these models and we added a feed-forward layer to classify a input sentence as `\textit{sustainable}' or `\textit{unsustainable}'.

\section{Experimental setup and evaluation}
\subsection{Evaluation metrics}

All runs were ranked based on mean rank and accuracy for the first sub-task and only accuracy for the second sub-task. The mean rank is the average of the ranks for all observations within each sample.

Accuracy determines how close the candidates' predictions are to their true labels:

\begin{equation}
    accuracy = \frac{1}{n_{samples}}\sum_{i=1}^{n_{samples}}1(\hat{y}_i = y_i),
\end{equation}

\noindent where $\hat{y}_i$ is the predicted value of the i-th sample and $y_i$ is the corresponding true value.

\subsection{Experimental evaluation}

In order to select the best pre-trained models for each sub-task, we split the training datasets into 70\% training and 30\% for development. Table~\ref{tb:training-split} shows the number of examples in the resulting training and development split for our analysis.

\begin{table}[tb]
\centering
\begin{tabular}{lcc}
\hline
\textbf{Sub-task}   & \textbf{\#Training} & \textbf{\#Dev} \\
\hline
ESG taxonomy extension & 452        & 195   \\
Sustainability classification & 1585       & 680   \\
\hline
\end{tabular}
\caption{Details of the split of the `Fortia ESG taxonomy' dataset to set our meta-parameters.}
\label{tb:training-split}
\end{table}

For the first sub-task, we selected the sentence BERT models: `\textit{bert-base-nli-mean-tokens}'\footnote{\url{https://huggingface.co/sentence-transformers/bert-base-nli-mean-tokens}}, `\textit{all-roberta-large-v1}'\footnote{\url{https://huggingface.co/sentence-transformers/all-roberta-large-v1}}, and `\textit{paraphrase-mpnet-base-v2}'\footnote{\url{https://huggingface.co/sentence-transformers/paraphrase-mpnet-base-v2}}. The first and second pre-trained SBERT models are based on the well-know BERT-based language models (BERT and RoBERTa language models, respectively). 
The third pre-trained model was trained on the paraphrase dataset where two paraphrases have close representation.
Table~\ref{tb:analysis-st1} shows the results for each pre-trained model. The `\textit{paraphrase-mpnet-base-v2}' achieved the best results for both metrics. We assume that the analysis of paraphrases is similar to the analysis of terms that share the same concept, which allowed this model to outperform the other models.

\begin{table}[tb]
\centering
\begin{tabular}{lcc}
\hline
\textbf{SBERT model} & \textbf{Mean rank} & \textbf{Accuracy}     \\
\hline
bert-base-nli-mean-tokens  & 1.502  & 0.764 \\
all-roberta-large-v1       & 1.461  & 0.779 \\
paraphrase-mpnet-base-v2   & \textbf{1.349}  & \textbf{0.810} \\
\hline
\end{tabular}
\caption{Results of our approach (Section~\ref{sc:approach1}) using different SBERT models for the first sub-task.}
\label{tb:analysis-st1}
\end{table}

For the second sub-task, we selected the BERT language models: DistilBERT~\cite{distilbert}, BERT, and RoBERTa. RoBERTa (Robustly Optimized BERT Pre-training Approach) is an extension of BERT with changes to the pre-training procedure~\cite{liu2019roberta}. They trained their model with bigger batches and over more data with long sentences. They also removed the next sentence prediction objective and dynamically changed the masking pattern applied to the training data. In this case, the RoBERTa language model outperformed the other models (Table~\ref{tb:analysis-st2}).

\begin{table}[]
\centering
\begin{tabular}{lc}
\hline
\textbf{BERT model}     & \textbf{Accuracy}     \\
\hline
distilbert-base-uncased & 0.906 \\
bert-base-uncased       & 0.921 \\
roberta-base            & \textbf{0.922} \\
\hline
\end{tabular}
\caption{Results of our approach (Section~\ref{sc:approach2}) using different BERT-based language models for the second sub-task.}
\label{tb:analysis-st2}
\end{table}

\subsection{Official results}

We submitted two runs for ESG taxonomy extension. The first run used the approach described in Section~\ref{sc:approach1} to train our model on the training data (Fortia ESG taxonomy). For the second run, we extended the Fortia ESG taxonomy with our in-house ESG taxonomy\footnote{No terms from our ESG taxonomy appear in the test data set published by the organizers.} and we used the same procedure to train the model. Our ESG taxonomy consists of a total of 65 terms spread across 22 concepts. For both runs, we used the pre-trained SBERT model `\textit{paraphrase-mpnet-base-v2}'. 

Official results for the first sub-task are listed in Table~\ref{tb:results-st1}. Both of our runs achieved the best results for mean rank and accuracy. In fact, our siamese model provided a better semantic representation of terms and outperformed the other approaches. The extension of the training data with our taxonomy enabled our model to better analyze the context of terms and their corresponding concepts and, consequently, improved the accuracy of $0.014$ points. 

\begin{table}[tb]
\centering
\begin{tabular}{lcc}
\hline
\textbf{Team name}                        & \textbf{Mean rank} & \textbf{Accuracy} \\
\hline
kaka\_1                                   & 1.441 & 0.745  \\
kaka\_2                                   & 1.670 & 0.662  \\
kaka\_3                                   & 1.545 & 0.752  \\
JETSONS\_1                                & 1.972 & 0.607  \\
LIPI\_subtask1\_1                         & 1.517 & 0.710  \\
LIPI\_subtask1\_2                         & 1.669 & 0.703  \\
TCSWITM\_1                                & 1.462 & 0.772  \\
TCSWITM\_2                                & 1.448 & 0.779  \\
vishleshak\_task1                         & 1.614 & 0.683  \\
Baseline1                                 & 2.276 & 0.462 \\
Baseline2                                 & 1.524 & 0.745 \\
\hline
ours\_wo\_extended\_data & 1.262 & 0.834  \\
\textbf{ours\_with\_extended\_data} & \textbf{1.255} & \textbf{0.848}  \\
\hline
\end{tabular}
\caption{Official results for the first sub-task. Our approaches are listed at the bottom of the table. The best results are in bold. Our model \textit{ours\_wo\_extended\_data} was trained on the original training data provided by the organizers and the version \textit{ours\_with\_extended\_data} was trained on the original data set combined with our taxonomy.}
\label{tb:results-st1}
\end{table}

We also submitted two runs for the second sub-task. The first run follows the same idea described in Section~\ref{sc:approach1} to represent the sentences by using SBERT. Then, the logistic regression classifies these sentence representations into only two classes: `\textit{sustainable}' and `\textit{unsustainable}'.
The second run uses the deep-learning model described in Section~\ref{sc:approach2}. Our model uses the pre-trained RoBERTa language model and two feed-forward layers to classify a sentence into `\textit{sustainable}' or `\textit{unsustainable}'. 

Official results for the second sub-task are listed in Table~\ref{tb:results-st2}.  Our runs achieved the fourth best result. 
The combination of fine-tuned RoBERTa language model and feed-forward layers outperformed both baselines as well as our run with SBERT and logistic regression. Our models performed well (over 92\% accuracy) and was ranked among the top 5 systems (0.19 points below the best-performing system).

\begin{table}[tb]
\centering
\begin{tabular}{lc}
\hline
\textbf{Team name}                         & \textbf{Accuracy} \\
\hline
kaka\_4                                    & 0.927 \\
\textbf{kaka\_2}                           & \textbf{0.946} \\
CompLx\_1                                  & 0.936 \\
FORMICA2\_1                                & 0.883 \\
FORMICA2\_2                                & 0.888 \\
LIPI\_1                                    & 0.922 \\
LIPI\_2                                    & 0.932 \\
TCSWITM\_1                                 & 0.873 \\
vishleshak\_task2                          & 0.912 \\
JETSONS\_1                                 & 0.927 \\
Baseline1                                  & 0.497 \\
Baseline2                                  & 0.819 \\
\hline
ours\_sbert\_logistic\_regression  & 0.907 \\
ours\_roberta\_with\_ffnn  & 0.927 \\
\hline
\end{tabular}
\caption{Official results for the second sub-task. Our approaches are listed at the bottom of the table. The best results are in bold.}
\label{tb:results-st2}
\end{table}

\section{Conclusion}

This paper described the joint effort of the L3i laboratory of the University of La Rochelle and the Trading Central Labs in the \textit{Shared Task FinSim-4} evaluation campaign for the task of ESG in financial documents. For this task, we developed BERT-based models. Our model based on siamese sentence analysis achieved the best results for the first sub-task. For the second sub-task, our approach based on the RoBERTa model got the fourth position.

\section*{Acknowledgments}
This work has been partially supported by the TERMITRAD (2020-2019-8510010) project funded by the Nouvelle-Aquitaine Region, France.

\bibliographystyle{named}
\bibliography{ijcai22}

\end{document}